\documentclass[runningheads]{llncs}

\usepackage[final,year=2026,ID=786]{accv}

\usepackage{accvabbrv}

\usepackage{graphicx}
\usepackage{booktabs}
\usepackage{amsmath}
\usepackage{amssymb}
\usepackage{array}
\usepackage{placeins}
\usepackage{float}
\usepackage[accsupp]{axessibility}

\usepackage[breaklinks,hidelinks]{hyperref}
\newcommand{\method}{MFilter}
\newcommand{\xfeat}{XFeat}
\newcommand{\md}{MegaDepth1500}

\begin{document}
\raggedbottom

\title{MFilter: Pair-Conditioned Keypoint Matchability for Lightweight Sparse Matching}
\titlerunning{MFilter for Lightweight Sparse Matching}

\author{
Xinpeng Zhu\protect\setcounter{footnote}{3}\thanks{These authors contributed equally.}
\and
Zhixiong Zhang\protect\footnotemark[4]
\and
Wentao Cheng
}

\authorrunning{X. Zhu, Z. Zhang, and W. Cheng}

\institute{
Beijing Normal-Hong Kong Baptist University, Zhuhai 519087, China\\
\email{
\{u430026305,u430026285\}@mail.bnbu.edu.cn,
wentaocheng@bnbu.edu.cn
}
}

\maketitle

\begin{abstract}
Lightweight sparse matching produces compact correspondences for geometric estimation, but descriptor similarity and per-image detector confidence do not fully capture reliability for a particular image pair.
We introduce \method{}, a pair-conditioned keypoint matchability filter for a fixed sparse candidate set.
A shared head predicts keypoint-side scores and symmetrically fuses them with descriptor similarity before ranking and selection, leaving feature extraction, MNN matching, and geometric estimation unchanged.
Controlled XFeat experiments show a modest aggregate pose-AUC gain over similarity filtering at the same retention ratio.
Additional experiments retrain the same formulation on SuperPoint, ALIKED, and DISK: retained-set inlier ratios improve, while pose gains depend on the backbone and strong handcrafted controls remain competitive.
Sequence-disjoint HPatches retraining improves correspondence statistics, while homography-AUC changes remain small and mixed.
The small scoring head adds modest measured latency, while the full pairwise-similarity workspace remains a memory limitation.
These results support a compact reliability module while highlighting a distinction between correspondence quality and downstream geometric utility.
\end{abstract}

\section{Introduction}
\label{sec:intro}

Local feature matching underpins two-view geometry, visual localization, 3D reconstruction, and visual measurement.
In many deployed or resource-constrained settings, the relevant question is not only which matcher is most accurate, but which matcher provides stable geometric evidence at low latency and memory cost.
Lightweight sparse pipelines remain useful for this reason: they produce compact correspondence sets that can be passed directly to robust pose solvers, and they avoid the cost of dense correlation or deep cross-image reasoning.

The same design choices that make lightweight sparse pipelines efficient can weaken their reliability estimates.
Many such methods rely on local CNN features, compact descriptors, and simple mutual-nearest-neighbor matching.
Limited receptive fields and per-image detector scores are effective for speed, but they give little direct evidence about whether a detected point is matchable for the current image pair.
Descriptor similarity measures local appearance agreement; it does not fully encode ambiguity with respect to competing matches, local spatial support, or downstream geometric utility.
Thus, single-image confidence and descriptor agreement are useful but incomplete proxies for pair-conditioned reliability.

One route to stronger matching is richer cross-image reasoning.
Learned sparse matchers such as SuperGlue and LightGlue model interactions among detected local features~\cite{Sarlin_2020_SuperGlue,Lindenberger_2023_LightGlue}, while detector-free and dense methods such as LoFTR and Efficient LoFTR reason over image feature maps~\cite{Sun_2021_CVPR,Wang_2024_EfficientLoFTR}.
These approaches are powerful, but they either replace the lightweight matching stage or add nontrivial computation.
We therefore study a complementary question: after a lightweight sparse matcher has already produced its candidate set, can a small pair-conditioned reliability module improve the candidate ordering without changing the rest of the pose pipeline?

We address this question with \method{}, a reliability filter after matching.
A shared head evaluates the two participating keypoints of each candidate using local evidence and cross-image ambiguity cues, then symmetrically modulates descriptor similarity before selection.
Our contribution is the keypoint-side, pair-conditioned formulation tied to a fixed MNN candidate set and a pipeline-derived supervision target.
The intervention is restricted to candidate ranking and selection, making its benefit and cost measurable within each sparse pipeline.

Our contributions are:
\begin{itemize}
  \item We formulate pair-conditioned keypoint matchability for fixed sparse candidates, using a shared head and symmetric score fusion.
  \item Matched-count controls and fusion variants characterize the ranking signal; stronger low-cost controls and backbone-specific retraining test its scope beyond XFeat.
  \item Cross-dataset and in-domain training experiments, together with latency and memory profiles, distinguish correspondence reliability from downstream geometric utility and deployment cost.
\end{itemize}

\section{Related Work}
\label{sec:related}

\paragraph{Sparse local features and descriptors.}
Classical sparse pipelines combine keypoint detection with local description and nearest-neighbor matching.
SIFT established the scale-invariant detect--describe paradigm~\cite{Lowe_2004_IJCV}, while ORB provided a faster binary alternative~\cite{Rublee_2011_ORB}.
Learning first improved patch descriptors through HardNet and SOSNet~\cite{Mishchuk_2017_HardNet,Tian_2019_SOSNet}, then increasingly learned detection and description jointly.
Representative systems include LIFT, SuperPoint, D2-Net, R2D2, DISK, Key.Net, and ASLFeat~\cite{Yi_2016_LIFT,DeTone_2018_SuperPoint,Dusmanu_2019_D2Net,Revaud_2019_R2D2,Tyszkiewicz_2020_DISK,BarrosoLaguna_2019_KeyNet,Luo_2020_ASLFeat}.
ALIKE and ALIKED further target efficient local feature extraction~\cite{Zhao_2022_ALIKE,Zhao_2023_ALIKED}, and \xfeat{} provides lightweight sparse and semi-dense operating modes~\cite{Potje_2024_CVPR}.
These methods primarily improve which features are detected or described. \method{} instead asks how to rank the correspondences produced by a fixed lightweight feature pipeline.

\paragraph{Learned sparse matchers.}
SuperGlue casts local feature matching as learned partial assignment over two keypoint sets~\cite{Sarlin_2020_SuperGlue}, and LightGlue reduces the cost through adaptive computation~\cite{Lindenberger_2023_LightGlue}.
GlueStick expands contextual matching to joint point and line structures~\cite{Pautrat_2023_GlueStick}, while AdaLAM enforces efficient local geometric consistency without global attention~\cite{Cavalli_2020_AdaLAM}.
These methods enrich or replace the matching rule.
In contrast, \method{} preserves the fixed \xfeat{} MNN candidates and adds only a reliability-based ordering after matching.

\paragraph{Detector-free, semi-dense, and dense matching.}
Detector-free systems LoFTR and Efficient LoFTR establish correspondences with cross-image feature interaction~\cite{Sun_2021_CVPR,Wang_2024_EfficientLoFTR}.
MatchFormer interleaves feature extraction and matching~\cite{Wang_2022_MatchFormer}, whereas ASpanFormer adapts the spatial extent of attention to the estimated matching uncertainty~\cite{Chen_2022_ASpanFormer}.
Dense methods such as DKM and RoMa predict image-wide correspondence fields and confidence~\cite{Edstedt_2023_DKM,Edstedt_2024_RoMa}.
This broader literature demonstrates the value of cross-image context, but generally departs from the sparse local-feature pipeline.
\method{} instead introduces a small pair-conditioned reliability signal while retaining the computational structure of sparse \xfeat{}.

\paragraph{Correspondence reliability and outlier filtering.}
Learned pruning methods infer inlier likelihood or geometric consensus from putative correspondences.
PointCN introduced context-normalized correspondence classification~\cite{Yi_2018_GoodCorrespondences}; OANet models order-aware context~\cite{Zhang_2019_OANet}; ACNe suppresses outliers during context normalization~\cite{Sun_2020_ACNe}; and CLNet progressively prunes candidates by local consensus~\cite{Zhao_2021_CLNet}.
NCMNet mines consistent neighborhoods~\cite{Liu_2023_NCMNet}, ConvMatch uses convolutional context aggregation~\cite{Zhang_2024_ConvMatch}, and BCLNet learns bilateral consensus~\cite{Miao_2024_BCLNet}.
Unlike correspondence-level pruning methods that infer consensus over a set of putative matches, \method{} uses pair-conditioned keypoint-side matchability to re-rank a fixed MNN candidate set under an otherwise unchanged sparse-\xfeat{} pose pipeline.
Similarity filtering at the same retention ratio controls for candidate count, while Lowe-ratio, margin, and detector-confidence rankings provide stronger tests against inexpensive alternative orderings.

\section{Method}
\label{sec:method}

\begin{figure}[t]
  \centering
  \includegraphics[width=\linewidth]{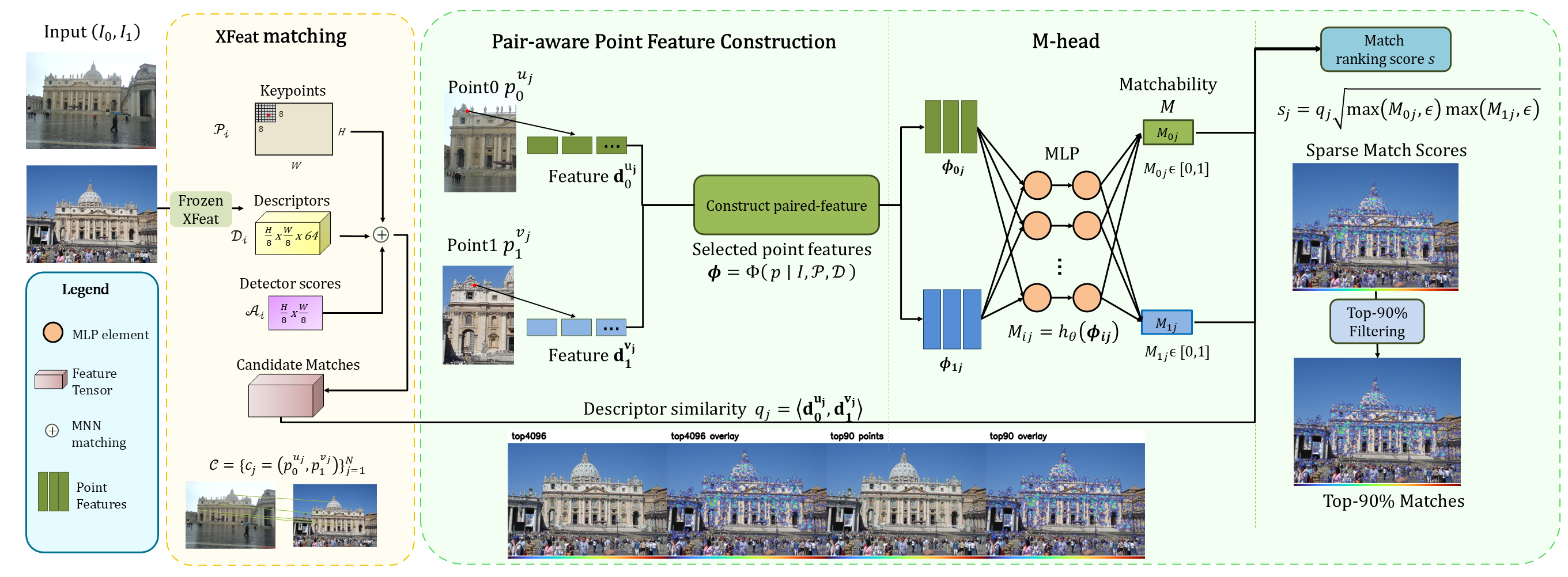}
  \caption{Overview of \method{}. Frozen sparse \xfeat{} provides keypoints, descriptors, detector scores, and MNN candidate matches. A shared matchability head evaluates pair-conditioned features for the two keypoints of each candidate. Their scores symmetrically gate descriptor similarity, after which the highest-ranked correspondences are passed to the unchanged pose solver. The schematic depicts the geometric-mean gate ($\alpha=0.5$); the validation-selected setting uses $\alpha=1.5$ in Eq.~\eqref{eq:mfilter-score}.}
  \label{fig:workflow}
\end{figure}

\subsection{Fixed Candidates and Pair-Conditioned Features}
\label{sec:pair-conditioned-matchability}
We describe the method using sparse \xfeat{}; the additional backbones use the same feature groups and head architecture with separately trained weights.
For images $(I_a,I_b)$, the frozen extractor returns $\mathcal F_x=\{(\mathbf p_x^k,a_x^k,\mathbf d_x^k)\}_{k=1}^{n_x}$, $x\in\{a,b\}$, comprising 2D keypoint locations, detector scores, and descriptors.
MNN matching produces $\mathcal C=\{c_j=(u_j,v_j,q_j)\}_{j=1}^{N}$, where $u_j,v_j$ index the two participating keypoints and $q_j=\langle\mathbf d_a^{u_j},\mathbf d_b^{v_j}\rangle$ is their descriptor similarity.
\method{} changes only the ranking and retained subset of $\mathcal C$ (Fig.~\ref{fig:workflow}).

For each detected keypoint, $\Phi(\mathbf p_x^k;\mathcal F_x,\mathcal F_{\bar x})\in\mathbb R^{31}$ describes confidence, position, spatial support, and opposite-image alternatives, where $\bar x$ denotes the other image.
The same construction is applied after swapping the image order; candidate indices then select the two keypoint-side vectors $\boldsymbol\phi_a^j$ and $\boldsymbol\phi_b^j$.
Table~\ref{tab:features} lists the 20 single-image and 11 cross-image components in feature order.
The cross-image components make the representation pair-conditioned without a learned global interaction network.

\begin{table}[t]
\caption{The 31 input components. ``Other'' denotes the best descriptor match in the opposite image.}
\label{tab:features}
\centering\small
\begin{tabular}{@{}>{\raggedright\arraybackslash}p{0.23\linewidth}c>{\raggedright\arraybackslash}p{0.63\linewidth}@{}}
\toprule
Group & Dim. & Components \\
\midrule
Order & 2 & Detector rank; normalized rank \\
Confidence & 3 & Raw, image-standardized, and image-mean-normalized score \\
Location & 3 & Normalized $x$, $y$, and center distance \\
Descriptor & 1 & Descriptor $\ell_2$ norm \\
Grid support & 4 & Normalized cell count; cell-score mean, maximum, standard deviation \\
Score context & 4 & Score/cell-mean; score/cell-maximum; global and cell percentiles \\
Distribution & 3 & Occupied-cell fraction; normalized grid-count and score entropies \\
Alternatives & 4 & Best and second-best similarities; their difference and ratio \\
Mutuality & 1 & Mutual best-assignment indicator \\
Other conf. & 3 & Rank, normalized rank, detector score \\
Other location & 3 & Normalized $x$, $y$, center distance \\
\bottomrule
\end{tabular}
\end{table}

Coordinates are divided by $(W-1,H-1)$, with clipped indices assigning points to an $8\times8$ grid. Descending detector ranks are divided by $\max(n-1,1)$; global and cell percentiles decrease from 1 to 0. Cell counts are divided by $n$. Score ratios use an additive $\epsilon=10^{-6}$ denominator, and entropies are normalized by the logarithm of their nonempty support. A scaler fitted only on training keypoints standardizes all components before the head.

\subsection{M-Label Supervision and Shared Head}
\label{sec:keypoint-matchability}
Let $z_j\in\{0,1\}$ indicate whether candidate $j$ is accepted by the frozen pipeline's RANSAC verifier.
A keypoint is positive if it participates in at least one verified MNN correspondence:
\begin{equation}
y_a^k=\mathbf1[\exists j:\ u_j=k\land z_j=1],\qquad
y_b^\ell=\mathbf1[\exists j:\ v_j=\ell\land z_j=1].
\label{eq:m-label}
\end{equation}
These M-labels provide pipeline-derived supervision for matchability under the chosen matcher and RANSAC verifier.

A shared head predicts the two keypoint-side scores for each candidate:
\begin{equation}
m_a^j=h_\theta(\overline{\boldsymbol\phi}_a^j),\qquad
m_b^j=h_\theta(\overline{\boldsymbol\phi}_b^j),\qquad m_a^j,m_b^j\in[0,1].
\label{eq:matchability-head}
\end{equation}
The overbar denotes training-set standardization. The head is a $31\!\rightarrow\!64\!\rightarrow\!64\!\rightarrow\!1$ MLP, with ReLU and dropout 0.1 in each hidden layer and a sigmoid output.
For labeled training keypoints $t\in\mathcal T$, class-balanced binary cross-entropy minimizes the average of $-w^+y_t\log m_t-(1-y_t)\log(1-m_t)$.
For the XFeat-specific \method{} head, the positive-class weight is $w^+=37.80$.

\subsection{Symmetric Fusion and Selection}
\label{sec:match-ranking}
For candidate $c_j$, we fuse descriptor similarity with its two keypoint-side scores:
\begin{equation}
s_j=q_j\left[\max(m_a^j,\epsilon)\max(m_b^j,\epsilon)\right]^\alpha.
\label{eq:mfilter-score}
\end{equation}
The shared head and symmetric product make this score invariant to swapping the images.
The exponent $\alpha$ controls the influence of matchability and $\epsilon=10^{-6}$ stabilizes the product.
The factorization exposes weak support at either participating keypoint and permits one-sided and matchability-only controls.

Sorting the $N$ candidates by $s_j$, we retain the $K=\min(N,\max(K_{\min},\lceil rN\rceil))$ highest-scoring candidates and pass their coordinates to the unchanged geometric solver.
For the selected $r=0.90$ and $K_{\min}=100$, small candidate sets can retain more than 90\%; if fewer than 100 are available, all are kept.
Selection depends on relative ranking rather than an absolute score threshold, and controls with the same retention ratio share this count rule.
No additional geometric verification is introduced at inference.

\subsection{Inference Complexity}
\label{sec:inference-complexity}
The current implementation forms an $n_a\times n_b$ descriptor-similarity matrix to retrieve the best and second-best alternatives for every keypoint.
For descriptor dimension $d$, this costs $O(n_an_bd)$ time and $O(n_an_b)$ memory.
Image-side statistics require global and grid-cell score sorting.
With fixed hidden width, shared-head evaluation is linear in $n_a+n_b$, fusion costs $O(N)$, and candidate sorting costs $O(N\log N)$.
The reported runtime includes feature construction and computation of the pairwise similarity matrix.
We therefore report both latency and peak CUDA allocation in Sec.~\ref{sec:runtime-analysis}.

\section{Experiments}
\label{sec:experiments}

\subsection{Implementation Details}
\label{sec:implementation-details}

\paragraph{Evaluation protocol.}
We evaluate relative pose on \md{} using AUC at $5^\circ$, $10^\circ$, and $20^\circ$, reported as percentage points together with their mean.
All controlled variants start from the same top-4096 sparse \xfeat{} detections, MNN candidates, and essential-matrix RANSAC with a 2.0-pixel threshold.
The \md{} test set contains 1,500 fixed image pairs.
HPatches uses 116 sequences and 580 pairs; ScanNet uses the official 1,500-pair protocol.
\paragraph{Feature specification.}
Table~\ref{tab:features} defines the complete 31-component input and its normalization. The same formulation is used for all evaluated backbones; each backbone has its own fitted scaler and head.

\paragraph{Matchability head and optimization target.}
The shared head is a $31\!\rightarrow\!64\!\rightarrow\!64\!\rightarrow\!1$ MLP.
Each hidden layer uses ReLU and dropout 0.1; a sigmoid converts the final logit to a matchability value.
Class-balanced binary cross-entropy uses positive weight $w^+=37.80$ because positive M-labels are sparse.
The training split contains 8.19 million keypoints from 1,000 outdoor pairs, with a positive rate of 2.58\%.
The validation split contains 1.64 million keypoints from 200 disjoint pairs, with a positive rate of 1.87\%.
Training uses Adam with learning rate $10^{-3}$, batch size 8192, and 20 epochs.
The epoch-18 checkpoint is selected by validation PR-AUC and reaches 0.955 ROC-AUC and 0.400 PR-AUC against the RANSAC-derived validation labels.
The main and cross-backbone experiments use seed 42; validation PR-AUC selects epochs 3, 4, and 2 for the SuperPoint, ALIKED, and DISK heads, respectively. ScanNet training uses seed 20260830. HPatches uses deterministic sequence-disjoint folds, each with its own training-only scaler, head, and checkpoint selection.

\paragraph{Data separation and model selection.}
Training M-labels are generated by running the frozen \xfeat{} MNN matcher and RANSAC on the training pairs.
The feature scaler, class weight, and MLP parameters are fitted only on these keypoints.
The same label rule is applied to the disjoint validation pairs, but validation data is used only for checkpoint selection and the independent sweep over $\alpha$ and $r$.
The \md{} test pairs are held out from feature standardization, head training, checkpoint selection, and operating-point selection.
The test pairs are used only for evaluation and the sensitivity analyses reported below.

\paragraph{Operating point and runtime.}
The independent validation sweep selects $\alpha=1.5$ and $r=0.90$; the minimum retained count is $K_{\min}=100$.
All other $\alpha$ and $r$ rows are reported as post-selection sensitivity analyses.
At least 100 candidates are kept whenever they are available, so a pair with fewer than 112 candidates can retain more than 90\%.
Runtime is measured from image-pair matching through pose RANSAC over the same 1,500-pair run, using explicit CUDA synchronization and 20 warm-up pairs.
\subsection{MegaDepth1500 Controlled Comparison}
\label{sec:megadepth-comparison}

Table~\ref{tab:megadepth-context} places \method{} in the broader matching landscape.

\begin{table}[t]
  \caption{Contextual comparison on \md{} (AUC in percentage points). Pipeline types and descriptor dimensions differ across rows. \xfeat{}* denotes the official semi-dense mode.}
  \label{tab:megadepth-context}
  \centering
  \resizebox{\linewidth}{!}{%
  \begin{tabular}{llccccc}
    \toprule
    Method & Pipeline & AUC@5 & AUC@10 & AUC@20 & Mean & Dim \\
    \midrule
    SuperPoint+LG & learned sparse & 66.30 & 79.01 & 87.75 & 77.69 & 256 \\
    DISK+LG & learned sparse & 61.34 & 74.48 & 83.95 & 73.26 & 128 \\
    ALIKED+LG & learned sparse & 65.97 & 78.69 & 87.38 & 77.35 & 64 \\
    SIFT+LG & learned sparse & 64.65 & 77.83 & 86.87 & 76.45 & 128 \\
    \xfeat{}* & lightweight semi-dense & 50.00 & 65.10 & 77.00 & 64.03 & 64 \\
    ALIKE & lightweight sparse & 50.20 & 61.80 & 70.80 & 60.93 & 64 \\
    \midrule
    \xfeat{} & lightweight sparse & 43.14 & 56.43 & 67.30 & 55.62 & 64 \\
    \xfeat{}+\method{} & sparse + filter & 44.18 & 57.55 & 68.23 & 56.65 & 64 \\
    \bottomrule
  \end{tabular}
  }
\end{table}

Table~\ref{tab:megadepth-controlled} provides the controlled comparison by fixing the extractor, MNN candidates, and solver.
Raw \xfeat{} keeps all candidates; descriptor-only filtering and \method{} share the retained count, isolating the effect of their ordering.

\begin{table}[t]
  \caption{Controlled sparse-\xfeat{} comparison on \md{} (AUC in percentage points). The last two rows retain the same number of candidates; bold marks the best value.}
  \label{tab:megadepth-controlled}
  \centering
  \resizebox{\linewidth}{!}{%
  \begin{tabular}{lclcccc}
    \toprule
    Method & Keep & Ranking score & AUC@5 & AUC@10 & AUC@20 & Mean \\
    \midrule
    \xfeat{} & 1.00 & $q_j$ & 43.14 & 56.43 & 67.30 & 55.62 \\
    \xfeat{}-sim & 0.90 & $q_j$ & 42.55 & 56.48 & 67.71 & 55.58 \\
    \textbf{\xfeat{}+\method{}} & 0.90 & $q_j(m_a^jm_b^j)^{\alpha}$ & \textbf{44.18} & \textbf{57.55} & \textbf{68.23} & \textbf{56.65} \\
    \bottomrule
  \end{tabular}
  }
\end{table}

This controlled sparse-\xfeat{} setting isolates the ranking signal: \method{} changes neither the detector, descriptor, MNN rule, descriptor dimensionality, nor pose solver.
Since descriptor-only filtering applies the same retained count, its proximity to the raw baseline is a direct control for simple match reduction.
The aggregate mean-AUC comparison favors \method{} over both raw \xfeat{} and the control with the same retention ratio.

\subsection{Cross-Backbone Evaluation and Stronger Low-Cost Controls}
\label{sec:backbones}
We train a separate head for each of SuperPoint, ALIKED, and DISK on 1,000 training and 200 validation pairs using the same 31-dimensional features.
The MegaDepth test set is reserved for evaluation.
All filtered variants use $r=0.90$, $K_{\min}=100$, and $\alpha=1.5$ for \method{}.
Within each backbone, controls share the same candidate set, retained count, and geometric solver.
They include descriptor similarity $q_j$, Lowe-style ambiguity $-\max(\rho_a,\rho_b)$, similarity margin $\min(\delta_a,\delta_b)$, and detector-confidence ranking $q_j\sqrt{a_a a_b}$, where $\rho$ is the nearest-to-second-nearest descriptor-distance ratio and $\delta$ is the best-minus-second-best descriptor similarity.
Table~\ref{tab:backbones} compares \method{} with the strongest reported low-cost control for each backbone.

\begin{table}[t]
\caption{MegaDepth1500 experiments with backbone-specific heads. Mean pose AUC and the MFilter-minus-control difference are percentages. Each row uses the strongest reported low-cost control for that backbone.}
\label{tab:backbones}
\centering\setlength{\tabcolsep}{5pt}\small
\begin{tabular}{llrrr}
\toprule
Backbone & Best low-cost control & Control & MFilter & Difference (\%) \\
\midrule
XFeat & Margin & \textbf{56.84} & 56.65 & $-0.19$ \\
SuperPoint & Margin & 55.32 & \textbf{56.02} & $+0.70$ \\
ALIKED & Lowe ratio & \textbf{69.17} & 68.54 & $-0.63$ \\
DISK & Margin & \textbf{65.12} & 64.72 & $-0.40$ \\
\bottomrule
\end{tabular}
\end{table}

\method{} exceeds the strongest reported low-cost control on SuperPoint.
Margin ranks higher on XFeat and DISK; Lowe ratio ranks higher on ALIKED.
Relative to similarity filtering at the same retention ratio, SuperPoint mean pose AUC improves by 1.00\%. Retained-set RANSAC inlier ratio improves by 0.40\%, 0.18\%, and 1.58\% on SuperPoint, ALIKED, and DISK, respectively, while their pose-AUC changes are $+1.00\%$, $-0.27\%$, and $+0.47\%$.
The mixed pose results also show that increased inlier ratio need not yield higher pose AUC.

\subsection{Runtime Analysis}
\label{sec:runtime-analysis}

Table~\ref{tab:runtime-breakdown} reports the canonical synchronized RTX 4090 profile. We average over 1,500 MegaDepth test pairs after 20 warm-up pairs.
Timing uses explicit CUDA synchronization around matcher calls and covers matching through pose estimation rather than an isolated head-only benchmark.
\method{} adds pair-conditioned feature construction, two evaluations of the shared head, fusion, and sorting.

\begin{table}[t]
\caption{Canonical synchronized RTX 4090 profile. Latency covers the end-to-end matching-through-pose pipeline; peak CUDA allocation includes the pairwise-similarity workspace.}
\label{tab:runtime-breakdown}
\centering\setlength{\tabcolsep}{5pt}
\begin{tabular}{lrrr}
\toprule
Pipeline & Latency (ms) & $\Delta$ latency (ms) & Peak CUDA allocation (MiB) \\
\midrule
XFeat & 42.94 & -- & 144.3 \\
XFeat + MFilter & 47.96 & $+5.02$ & 464.4 \\
\bottomrule
\end{tabular}
\end{table}

The head contains 6,273 parameters (24.5 KiB in FP32) and adds 5.02 ms in the synchronized end-to-end profile.
The current full-similarity implementation adds 320.1 MiB to peak CUDA allocation.
The pairwise-similarity workspace therefore remains the main memory cost of the current implementation.

\subsection{Reliability and Filtering Analysis}
\label{sec:reliability-filtering-analysis}

We next analyze how the learned score changes the retained candidate set.
On the 1.64-million-keypoint validation set, the M-label positive rate is effectively zero in the lowest seven equal-count score bins and rises to 0.73\%, 3.07\%, and 14.87\% in the top three.
The score bins therefore track consistency with the RANSAC-derived M-labels.

\begin{table}[t]
  \caption{Score-fusion sensitivity on \md{} at $r=0.90$ (AUC and inlier ratio in percentages). All variants use the same candidates. The validation-selected $\alpha=1.5$ row gives the highest mean AUC among these variants.}
  \label{tab:score-fusion-ablation}
  \centering
  \small
  \setlength{\tabcolsep}{4pt}
  \begin{tabular}{lccccc}
    \toprule
    Ranking score & AUC@5 & AUC@10 & AUC@20 & Mean & Inlier ratio \\
    \midrule
    $q_j$ & 42.55 & 56.48 & 67.71 & 55.58 & 55.89 \\
    $(m_a^jm_b^j)^{0.5}$ & 43.93 & 57.17 & 67.99 & 56.36 & 56.45 \\
    $q_j(m_a^jm_b^j)^{0.5}$ & 43.67 & 57.38 & 67.94 & 56.33 & 56.41 \\
    $q_jm_a^j$ & 43.51 & 56.83 & 67.55 & 55.97 & 56.38 \\
    $q_jm_b^j$ & 42.71 & 56.48 & 67.49 & 55.56 & 56.39 \\
    $q_jm_a^jm_b^j$ & 43.92 & 57.43 & 68.22 & 56.53 & 56.41 \\
    $q_j(m_a^j+m_b^j)/2$ & 43.46 & 57.05 & 67.98 & 56.17 & 56.42 \\
    $q_j(m_a^jm_b^j)^{1.5}$ & \textbf{44.18} & \textbf{57.55} & \textbf{68.23} & \textbf{56.65} & \textbf{56.48} \\
    \bottomrule
  \end{tabular}
\end{table}

Table~\ref{tab:score-fusion-ablation} shows that ranking by matchability alone gives higher mean AUC than ranking by descriptor similarity alone, indicating information beyond descriptor similarity.
Combining descriptor evidence with the two keypoint-side scores gives the strongest mean result and exceeds either one-sided gate.
The learned score therefore changes the composition of the retained candidate set.

\begin{figure}[t]
  \centering
  \begin{minipage}[t]{0.58\linewidth}
    \centering
    \includegraphics[width=\linewidth]{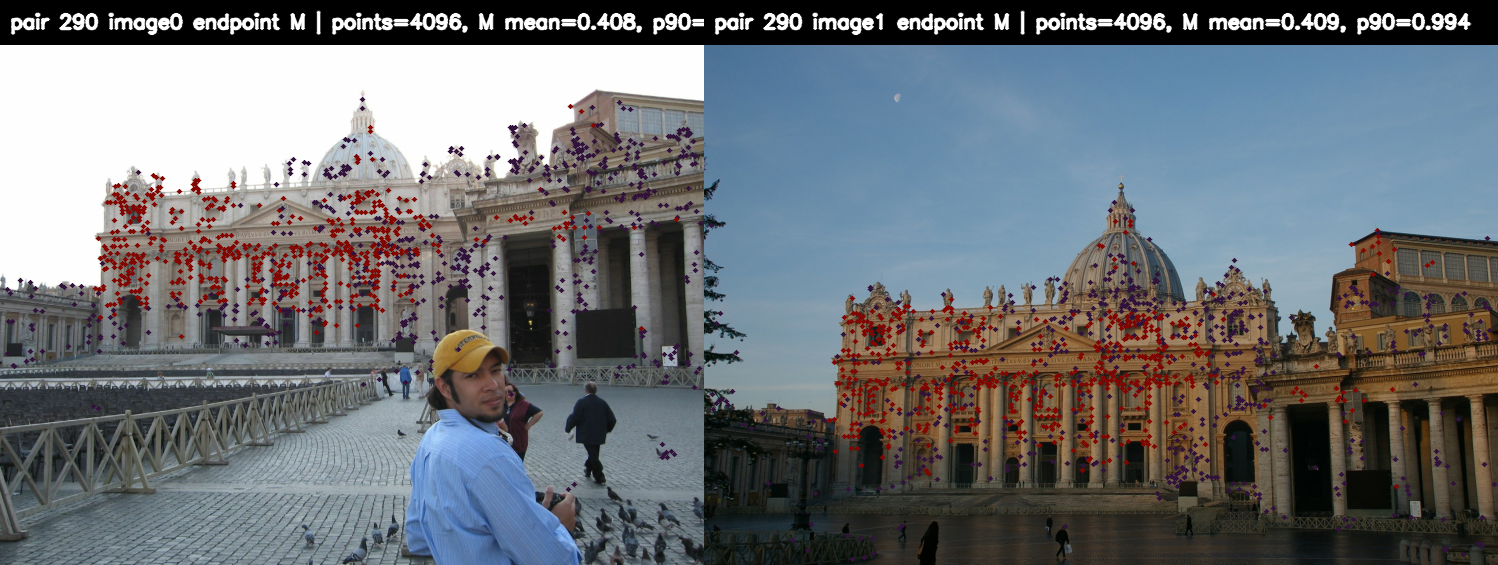}
    \vspace{1mm}
    {\small (a) Keypoint matchability.}
  \end{minipage}\hfill
  \begin{minipage}[t]{0.38\linewidth}
    \centering
    \includegraphics[width=\linewidth]{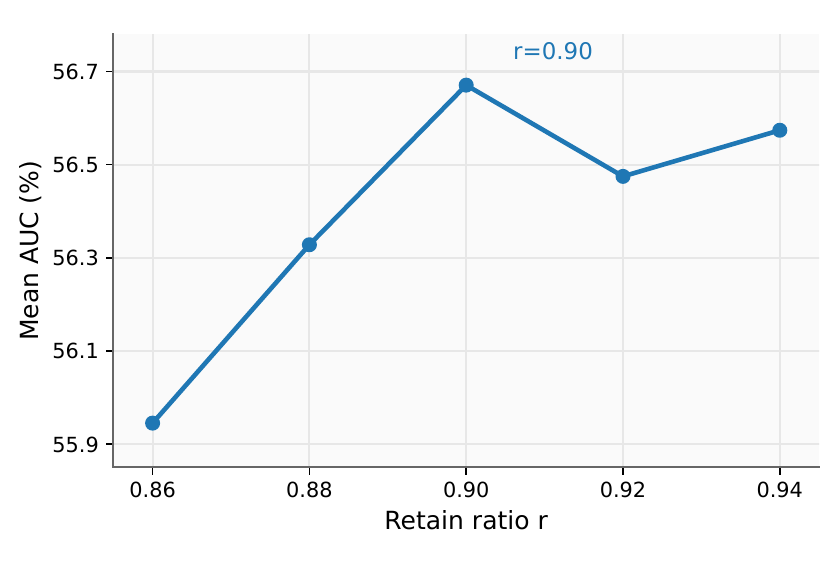}
    \vspace{1mm}
    {\small (b) Retain-ratio sweep.}
  \end{minipage}
  \caption{Reliability and filtering analysis. (a) Each \xfeat{} keypoint is colored by the shared matchability head; warmer colors indicate larger predicted matchability. (b) Ranking with $q_j(m_a^jm_b^j)^{1.5}$ removes the lowest-scoring tail as $r$ decreases; the highlighted operating point is selected on validation data.}
  \label{fig:matchability-analysis}
\end{figure}

\begin{figure}[t]
  \centering
  \includegraphics[width=\linewidth]{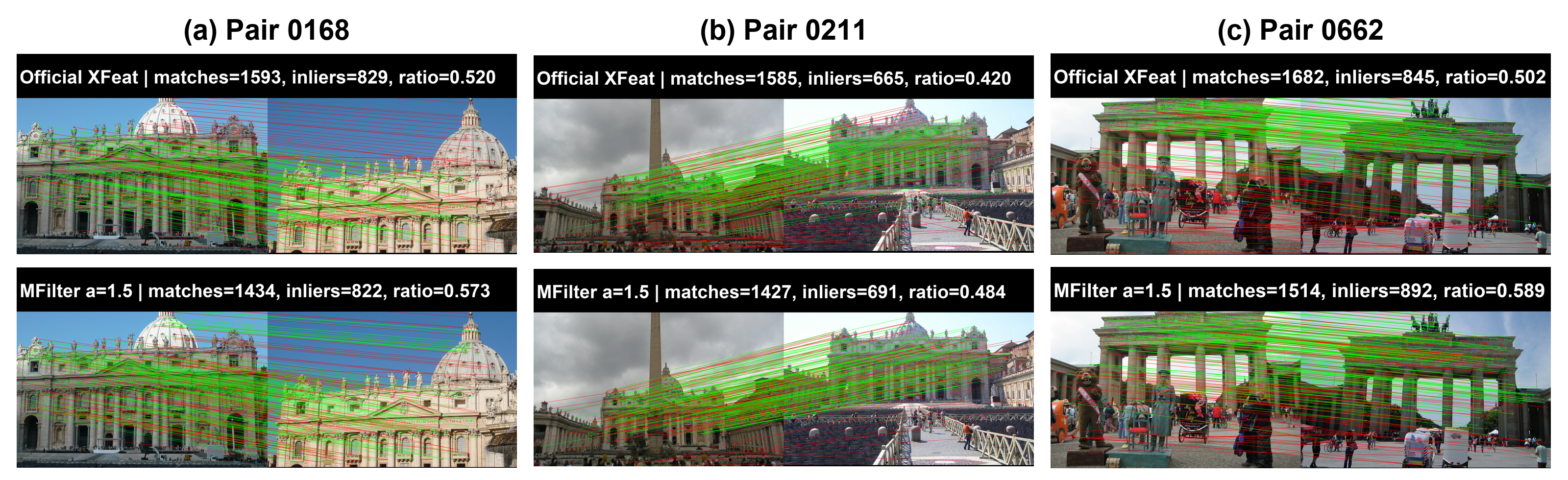}
  \caption{Selected qualitative examples on \md{}. Each panel compares the official sparse baseline with \method{} on the same image pair. The examples illustrate that pair-conditioned matchability tends to remove low-scoring correspondences while preserving spatially distributed support for pose estimation.}
  \label{fig:qualitative-three-pairs}
\end{figure}

Across \md{}, \method{} yields fewer average RANSAC inliers than raw \xfeat{} (830.84 vs.\ 852.22) while achieving higher aggregate pose AUC.
Together with the non-monotonic retain-ratio sweep in Fig.~\ref{fig:matchability-analysis}(b) and the descriptor-only control, this shows that performance depends on the composition of the retained set rather than pruning strength or inlier count alone.
Figure~\ref{fig:qualitative-three-pairs} illustrates this behavior: low-scoring local clusters are reduced while spatially distributed support is preserved.

\subsection{Cross-Dataset Evaluation}
\label{sec:cross-dataset-evaluation}

We evaluate cross-dataset transfer by applying the outdoor-trained head and the validation-selected operating point to HPatches and ScanNet.
Additional exponent settings are reported as sensitivity analyses.
This experiment probes whether the learned matchability signal transfers across different geometry and evaluation metrics.

\begin{table}[t]
  \caption{HPatches transfer results over 116 sequences and 580 pairs. Values are percentage points except counts and runtime. All filtered variants share the same $r=0.90$ retain rule. Panel A reports geometry and precision; panel B reports correspondence counts and cost.}
  \label{tab:hpatches}
  \centering
  \small\setlength{\tabcolsep}{4pt}
  \begin{tabular}{lrrrr}
    \toprule
    \multicolumn{5}{l}{A. Homography and matching precision} \\
    Ranking & H-AUC@3 & H-AUC@5 & H-AUC@10 & MMA@3 \\
    \midrule
    Raw XFeat & 36.74 & 51.45 & 68.88 & 55.10 \\
    Similarity & 38.12 & \textbf{52.49} & \textbf{69.26} & 58.46 \\
    MFilter ($\alpha=0.5$) & \textbf{38.55} & 52.28 & 69.00 & 58.80 \\
    MFilter ($\alpha=1.0$) & 37.81 & 51.57 & 68.63 & \textbf{58.83} \\
    MFilter ($\alpha=1.5$) & 38.08 & 52.16 & 69.21 & \textbf{58.83} \\
    \bottomrule
  \end{tabular}
  \par\medskip
  \begin{tabular}{lrrrr}
    \toprule
    \multicolumn{5}{l}{B. Retained correspondences and runtime} \\
    Ranking & Matches & Inliers & Inlier ratio & Runtime (ms) \\
    \midrule
    Raw XFeat & 1990.83 & \textbf{1239.38} & 62.25 & \textbf{20.11} \\
    Similarity & 1792.18 & 1180.98 & 65.90 & 21.42 \\
    MFilter ($\alpha=0.5$) & 1792.18 & 1186.48 & 66.20 & 33.17 \\
    MFilter ($\alpha=1.0$) & 1792.18 & 1187.65 & \textbf{66.27} & 33.50 \\
    MFilter ($\alpha=1.5$) & 1792.18 & 1187.56 & 66.26 & 33.75 \\
    \bottomrule
  \end{tabular}
\end{table}

On HPatches (Table~\ref{tab:hpatches}), the fixed $\alpha=1.5$ variant improves MMA@3 and inlier ratio over descriptor-similarity filtering at the same retention ratio, but its H-AUC is lower at all three thresholds.
The additional exponent settings reveal metric-dependent sensitivity around the selected operating point.
On HPatches, runtime rises from 20.11 to 33.75 ms, indicating dataset-dependent runtime overhead.

\begin{table}[t]
  \caption{ScanNet1500 transfer results using the official evaluator and a PoseLib threshold of 2.0 (AUC in percentage points).}
  \label{tab:scannet}
  \centering
  \resizebox{\linewidth}{!}{%
  \begin{tabular}{llccccc}
    \toprule
    Method & Score & Keep & AUC@5 & AUC@10 & AUC@20 & Mean \\
    \midrule
    \xfeat{} & $q_j$ & 1.00 & 14.01 & 28.55 & 43.84 & 28.80 \\
    \xfeat{}-sim & $q_j$ & 0.90 & 14.19 & 29.04 & 44.30 & 29.18 \\
    \xfeat{}+\method{} ($\alpha=0.5$) & $q_j(m_am_b)^{0.5}$ & 0.90 & 14.40 & 29.10 & 44.37 & 29.29 \\
    \xfeat{}+\method{} ($\alpha=1.0$) & $q_jm_am_b$ & 0.90 & 14.23 & 28.96 & 44.16 & 29.12 \\
    \xfeat{}+\method{} ($\alpha=1.5$) & $q_j(m_am_b)^{1.5}$ & 0.90 & \textbf{14.76} & \textbf{29.61} & \textbf{44.54} & \textbf{29.64} \\
    \bottomrule
  \end{tabular}
  }
\end{table}

At the fixed $\alpha=1.5$ operating point, ScanNet mean AUC increases from 28.80 to 29.64 (Table~\ref{tab:scannet}), with improvements at all three thresholds.
Among filtered variants, the validation-selected $\alpha=1.5$ setting achieves the highest AUC at all three thresholds.

\subsection{Dataset-Specific Training}
\label{sec:domain-training}
To examine whether domain shift accounts for limited transfer, we additionally train within HPatches while retaining the fixed filtering parameters.
For HPatches, sequence-level five-fold out-of-fold (OOF) evaluation covers 116 sequences and 580 pairs.
Each test sequence is excluded from training, validation, scaler fitting, and checkpoint selection.
Table~\ref{tab:domain-training} compares the outdoor-trained head and sequence-disjoint in-domain models.

\begin{table}[t]
\caption{HPatches sequence-disjoint five-fold OOF training at $\alpha=1.5$, $r=0.90$, and $K_{\min}=100$. H-AUC, MMA, and inlier ratio (IR) are percentages.}
\label{tab:domain-training}
\centering\setlength{\tabcolsep}{5pt}\small
\begin{tabular}{lrrrrrr}
\toprule
\multicolumn{7}{l}{HPatches, sequence-disjoint five-fold OOF} \\
Training & H-AUC@3 & @5 & @10 & MMA@3 & IR & Inliers \\
\midrule
Outdoor & 38.08 & 52.16 & 69.21 & 58.83 & 66.26 & 1187.56 \\
HPatches OOF & 38.07 & 52.07 & 69.33 & 59.21 & 66.70 & 1195.34 \\
\bottomrule
\end{tabular}
\end{table}

On HPatches, in-domain training improves MMA@3 by 0.38\%. Average inliers increase by 7.78, and inlier ratio increases by 0.43\%.
The H-AUC changes are small and mixed.

For ScanNet, 1,000 training and 200 selection pairs are drawn from independent scenes, while ScanNet1500 is reserved for evaluation.
Under the official PoseLib 2.0 protocol and a shared evaluation environment, the MegaDepth-trained head obtains 14.76/29.61/44.54 (mean 29.64), while the ScanNet-trained head obtains 14.55/29.27/44.55 (mean 29.45).
In-domain training does not improve mean AUC in this comparison.

\FloatBarrier
\subsection{Scope and Limitations}
\label{sec:limitations}
\method{} estimates pipeline-conditioned matchability because its M-labels are generated by the base matcher and RANSAC verifier.
Backbone-specific retraining extends the formulation to other sparse feature pipelines, while the cross-dataset experiments evaluate a fixed outdoor-trained head.
Downstream geometric gains vary across backbones and datasets, and handcrafted rankings remain competitive.
Inlier ratio captures correspondence purity, whereas spatial coverage and geometric leverage remain unmeasured.
The current full pairwise-similarity workspace also limits memory efficiency despite the small head.

\FloatBarrier
\section{Conclusion}
\label{sec:conclusion}
\method{} models pair-conditioned keypoint matchability with a shared keypoint-side head and symmetrically re-ranks fixed sparse correspondences.
The XFeat experiments support a modest aggregate benefit over the similarity control at the same retention ratio.
Across stronger controls and backbone-specific retraining, correspondence purity improves on the additional backbones, while pose gains depend on the backbone and strong handcrafted scores remain competitive.
Cross-dataset evaluation and HPatches OOF training further distinguish correspondence quality from downstream geometric utility.
The resulting contribution is a reusable reliability formulation with measured latency and workspace costs, whose downstream geometric benefits remain pipeline-dependent.

\section{Acknowledgement}
This work was supported by the National Natural Science Foundation of China (Grant No.~62306154).

\FloatBarrier
\clearpage
\bibliographystyle{splncs04}
\bibliography{main}
\end{document}